\documentclass[letterpaper, 10 pt, conference]{ieeeconf}  

\IEEEoverridecommandlockouts                              

\title{\LARGE \bf
SiCP: Simultaneous Individual and Cooperative Perception for 3D Object Detection in Connected and Automated Vehicles
}

\author{Deyuan Qu$^{1}$, Qi Chen$^{2}$, Tianyu Bai$^{1}$, Hongsheng Lu$^{2}$,
Heng Fan$^{1}$, Hao Zhang$^{3}$, Song Fu$^{1}$, Qing Yang$^{1}$\\
\thanks{$^{1}$University of North Texas, Denton, TX, USA}%
\thanks{$^{2}$Toyota InfoTech Labs, Mountain View, CA, USA}%
\thanks{$^{3}$University of Massachusetts Amherst, Amherst, MA, USA}%
}

\usepackage{hyperref}
\usepackage{graphicx}
\usepackage{amsmath}
\usepackage{amssymb}
\newcommand{\ignore}[1]{}
\usepackage{arydshln}
\usepackage{colortbl}
\usepackage{xcolor}
\usepackage{booktabs}
\usepackage{tabularx}
\usepackage{multirow} 
\usepackage{multicol}
\usepackage{subcaption}
\usepackage{epstopdf}
\definecolor{lightpink}{RGB}{255,210,220}
\definecolor{lightgreen}{RGB}{200,255,200}
\definecolor{lightblue}{RGB}{173,216,230}
\begin{document}

\maketitle
\thispagestyle{empty}
\pagestyle{empty}

\begin{abstract}
Cooperative perception for connected and automated vehicles is traditionally achieved through the fusion of feature maps from two or more vehicles. However, the absence of feature maps shared from other vehicles can lead to a significant decline in 3D object detection performance for cooperative perception models compared to standalone 3D detection models. This drawback impedes the adoption of cooperative perception as vehicle resources are often insufficient to concurrently employ two perception models. To tackle this issue, we present Simultaneous Individual and Cooperative Perception (SiCP), a generic framework that supports a wide range of the state-of-the-art standalone perception backbones and enhances them with a novel Dual-Perception Network (DP-Net) designed to facilitate both individual and cooperative perception. In addition to its lightweight nature with only 0.13M parameters, DP-Net is robust and retains crucial gradient information during feature map fusion. As demonstrated in a comprehensive evaluation on the V2V4Real and OPV2V datasets, thanks to DP-Net, SiCP surpasses state-of-the-art cooperative perception solutions while preserving the performance of standalone perception solutions. 
The source code can be found at \href{https://github.com/DarrenQu/SiCP}{https://github.com/DarrenQu/SiCP}.
\end{abstract}

\section{Introduction}
\begin{figure}[!htp]
  \begin{center}
  \includegraphics[width=3.2in, height=2.8in]{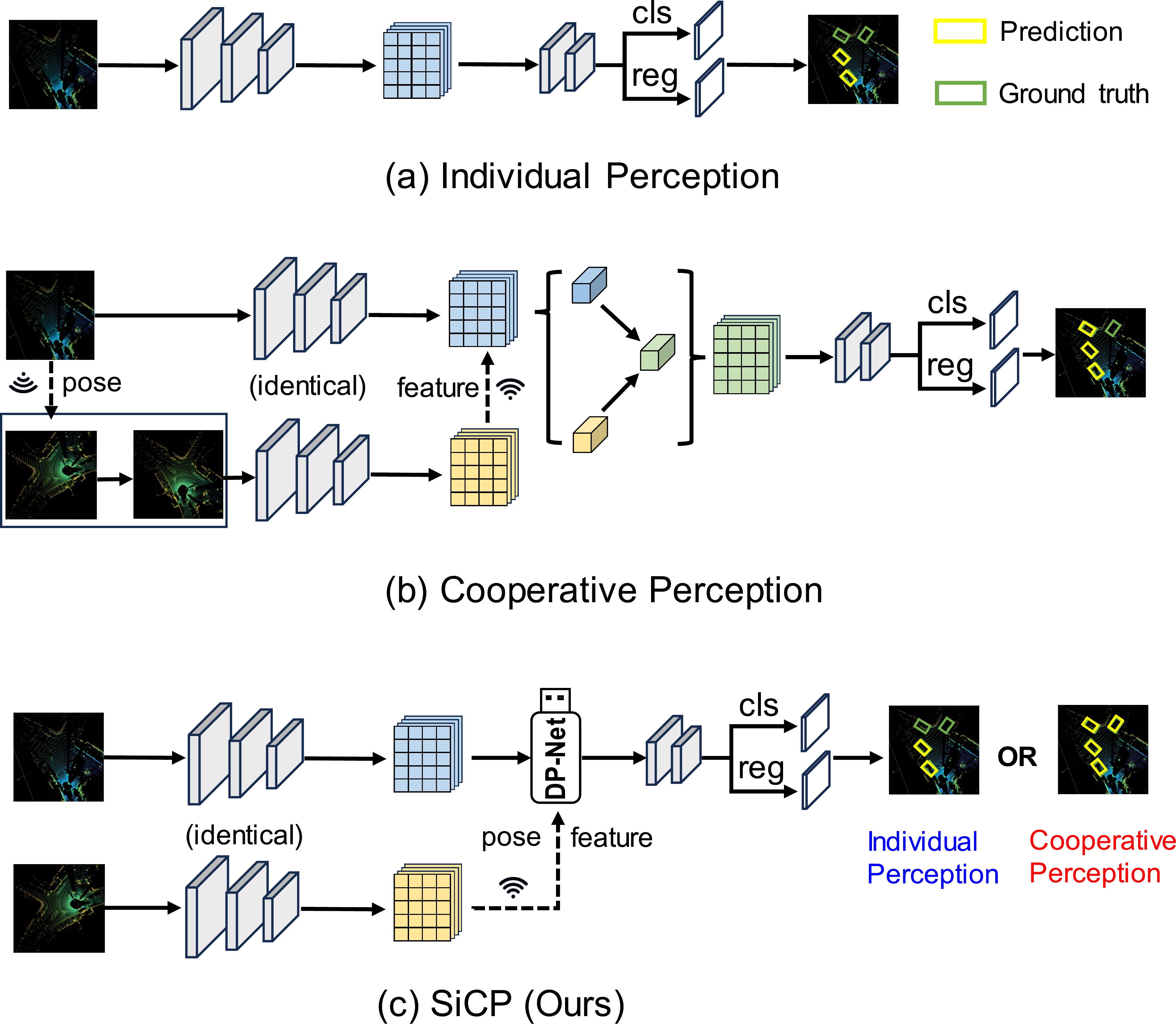}\\
   \caption{Different approaches to 3D perception. In (a), individual perception uses local sensor data for object detection. Cooperative perception, shown in (b), combines data from various vehicles to enhance the ego-vehicle's perception. Simultaneous Individual and Cooperative Perception (SiCP), as depicted in (c), supports both functionalities simultaneously.}
    \label{fig:idea}
    \vspace{-8mm}
  \end{center}
\end{figure}

Automated vehicles rely on standalone perception models to detect and comprehend 3D objects in their surrounding environment through individual sensors.
Cooperative perception, on the other hand, allows multiple vehicles to collaborate, enhancing their collective environmental awareness. 
Traditionally, these two modes of perception were studied in isolation, as shown in Figure~\ref{fig:idea} (a) and (b), neglecting the synergies between them. 
The significance of concurrently addressing these two facets was undervalued. 
In the realm of cooperative perception, researchers commonly adapt solutions from individual perception to collaborative settings. 
This adaptation involves expanding popular standalone 3D detection models, such as PointPillars~\cite{lang2019pointpillars}, SECOND~\cite{yan2018second} and VoxelNet~\cite{zhou2018voxelnet}, by fusing feature maps to become a cooperative perception models like F-Cooper~\cite{chen2019f}, AttFuse~\cite{xu2022opv2v}, V2X-ViT~\cite{xu2022v2x} and CoBEVT~\cite{xu2022cobevt}.  
With the exception of F-Cooper, most alternative methods require the sender to modify its local feature data to match the receiver's viewpoint prior to transmission, which proves to be impractical. 
The impracticality arises due to the potential existence of multiple receivers, and executing numerous transformations and transmissions of local feature maps introduces substantial computational and network overhead,
making the entire process excessively costly.
%
%
More importantly, the core problem with these cooperative perception models is their inability to effectively handle individual perception tasks. This concern is critical because, in real-world scenarios, there may not always be a sender vehicle available to share perception information. In such cases, ego vehicle must rely solely on its own perception capabilities to understand its environment. Our experiments indicate that cooperative perception models, as presented in Figure~\ref{fig:idea} (b), significantly lag behind the standalone perception model, as shown in Figure~\ref{fig:idea} (a), in performance on individual perception tasks.

This performance gap highlights the limitations of cooperative perception models in object detection scenarios where data is not shared between vehicles. 
A fundamental challenge arises from the divergence in features extracted from data of individual vehicles and multiple cooperating vehicles. 
The impact of these disparities remains poorly understood. 
To bridge this gap and pave the way for more robust and accurate automated vehicle perception systems, we explore the possibility of combining individual and cooperative perception to develop a unified framework that seamlessly integrates both individual and cooperative perceptions for connected and automated vehicles.
\subsection{Proposed Solution}
We present a pioneering perception framework, called SiCP (Simultaneous Individual and Cooperative Perception), to handle both individual and cooperative perceptions simultaneously, as illustrated in Figure~\ref{fig:idea} (c).
The SiCP architecture is composed of three key components: a feature extractor, a feature processor, and a detection head.
Initially, each vehicle utilizes an identical feature extractor to generate features suitable for both individual perception and cooperative perception tasks.
We devise a novel feature processor, named the Dual-Perception Network (DP-Net), to proficiently manage local features for individual perception and integrate fused features from neighboring vehicles for cooperative perception.
In situations where features from neighboring vehicles are unavailable, DP-Net relies on the ego vehicle's local feature map for individual perception tasks.
When such features are accessible, they are first transformed to the ego vehicle's perspective and then fused with ego vehicle's local feature map.
The DP-Net effectively merges Bird’s Eye View (BEV) feature maps by concatenating and condensing them into a single-channel feature map.
This condensed map undergoes further processing through two convolutional layers, resulting in a weighted map.
This weighted map plays a crucial role in adjusting the ego vehicle’s local feature map, while its complementary counterpart modifies the feature map received from other vehicles.
The adjusted feature maps are then concatenated and reshaped to the desired output size, successfully integrating information from neighboring vehicles in cooperative perception scenarios.
In terms of practical implementation, we embrace an approach akin to First-Come-First-Serve (FCFS), wherein the feature map of the ego vehicle is fused with those of the initially received neighbors, eliminating the necessity to await additional features before initiating the fusion process.
The resulting fused feature set is fed into the detection head, where it undergoes processing to generate classification and regression results for cooperative perception tasks. In cases where no additional features are received, the original local feature map is processed by the same detection head to complete individual perception tasks.
This approach ensures seamless integration of individual and cooperative perception, thereby improving overall efficiency and accuracy of the connected and automated vehicle perception system.
\subsection{Main Contributions}
The contributions of this work are as follows:
\begin{itemize}





    
    \item For the first time, we recognize the significance of employing a single 3D object detection network for simultaneous individual and cooperative perception within connected and automated vehicles. We present the SiCP framework as a novel solution to fill this gap.
    \item The proposed DP-Net is an innovative Plug-and-Play module as it can be seamlessly integrated into other standalone 3D detection models, enabling simultaneous individual and cooperative perception.
    \item The proposed DP-Net is also a lightweight component, comprising just 0.13M parameters, representing a mere 1.7\% increase from the standalone 3D perception model~\cite{lang2019pointpillars}.
    \item The proposed DP-Net exhibits robustness in addressing alignment errors caused by asynchronous communication and inaccurate localization between vehicles.

\end{itemize}

\section{Related Works}
\label{sec:literature_review}

\subsection{Individual Perception}
LiDAR-based 3D object detection plays a crucial role in the perception system of automated vehicles, effectively aiding in determining the size, position, and category of nearby 3D objects.
Currently, standalone 3D object detection models fall into two main categories: point-based and voxel-based. 
Point-based models, such as those proposed in~\cite{qi2017pointnet, shi2019pointrcnn, qi2017pointnet++}, directly process unstructured point clouds, extracting features directly from the raw data.
On the other hand, voxel-based methods, as exemplified by~\cite{zhou2018voxelnet, yan2018second, lang2019pointpillars}, transform point clouds into structured voxel or pillar formations. 
These methods adeptly balance computational performance with capturing essential spatial details.
Despite their strengths, both point-based and voxel-based methods face limitations in perception range and accuracy due to sensor constraints and the complexity of real-world road conditions. 
To address these challenges, there is a growing trend towards cooperative perception, which involves combining data from multiple vehicles, enhancing detection capabilities and overcoming individual perception limitations.

\subsection{Cooperative Perception}
Cooperative perception solutions for connected and automated vehicles classified into \textit{early fusion}~\cite{chen2019cooper, arnold2020cooperative}, \textit{deep fusion}~\cite{chen2019f,xu2022opv2v,xu2022v2x, xu2022cobevt, hu2022where2comm,lu2023robust,guo2021coff,vnet,li2021learning,xiang2023hm,ma2024macp,yu2022dair,dhakal2023sniffer,newdhakal2023sniffer,dhakal2023virtualpainting}, and \textit{late fusion}.
Among these, the {deep fusion} strikes a balance between bandwidth and detection performance, making it widely embraced in the literature.

\textbf{Activation function based deep fusion.} Initially introduced in~\cite{chen2019f}, F-Cooper employs the \textit{maxout} operation for feature map fusion.
CoFF~\cite{guo2021coff} enhances F-Cooper by incorporating feature enhancement techniques. 
Despite advancements, these solutions grapple with challenges stemming from heterogeneity of feature maps originating from different vehicles.
Even when focusing on the same region, perceptual differences can result in significantly varied features.

\textbf{Attention based deep fusion.} AttFuse~\cite{xu2022opv2v} incorporates a self-attention operation for feature fusion. 
Despite its effectiveness, the solution overlooks nearby features, missing out on crucial information locality.
V2X-ViT~\cite{xu2022v2x} introduces a unified transformer architecture for heterogeneous multi-agent perception, while CoBEVT~\cite{xu2022cobevt} presents a generic transformer-based framework.
\begin{figure*}[htbp]
    \centerline{
    \includegraphics[width=5.5in, height=2.2in]{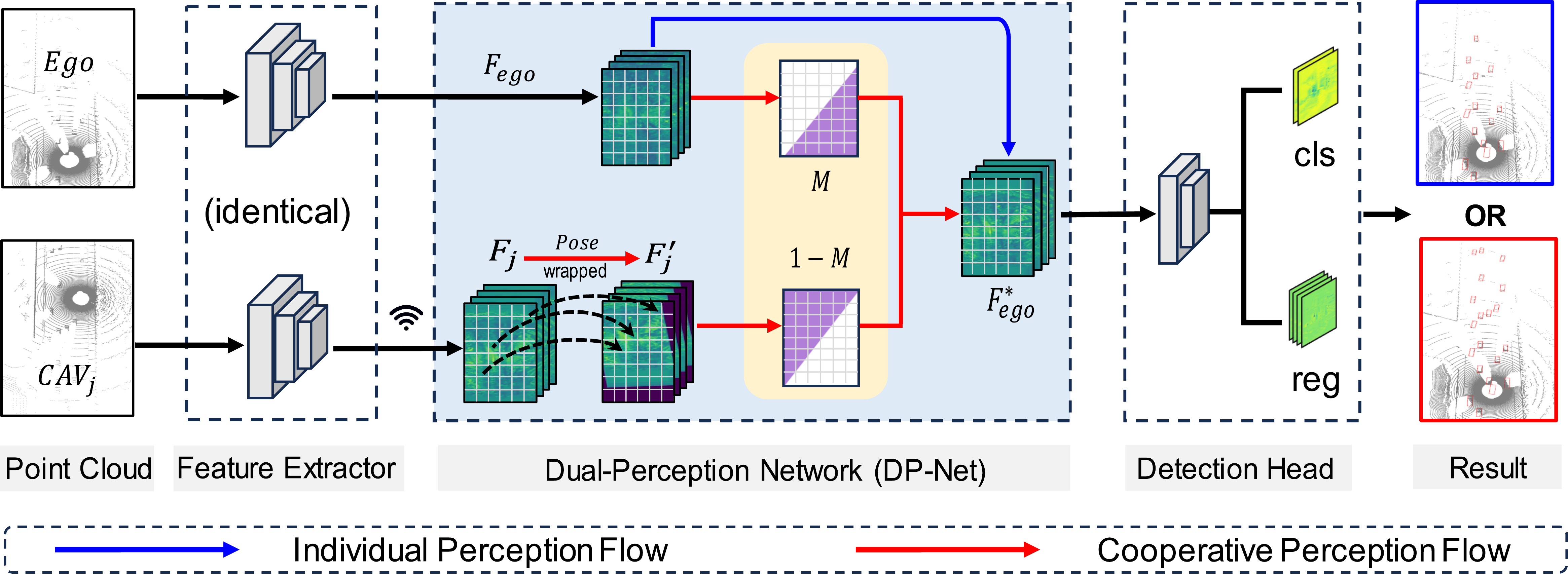}\\
    }
    \caption{An overview of the SiCP architecture showcases its components: a feature extractor, a feature processor (DP-Net), and a detection head. All vehicles have identical feature extractors producing fusible features. The feature processor manages local features $F_{ego}$ for individual perception and fused features $F^*_{ego}$ for cooperative perception. Features from other vehicles (e.g., $F_j$) are transformed to the ego vehicle's perspective and then performs a complementary fusion with the local features of the ego vehicle. The resulting feature $F^*_{ego}$ is then processed by the detection head to generate classification and regression results for either individual or cooperative perceptions.}
    \label{fig:SiCP}
    \vspace{-3mm}
\end{figure*}
While these solutions consider self-attention relations among all points in feature maps, making them computationally intensive, they are less focused on specific regions.
Other models adopt diverse strategies for feature fusion from various perspectives. 
For instance, Where2comm~\cite{hu2022where2comm} introduces a spatial confidence map to capture the spatial diversity of perceptual data, effectively minimizing communication bandwidth. 
%
%
%
Additionally, CoAlign~\cite{lu2023robust} introduces a novel hybrid collaboration framework designed to address pose errors. 
Furthermore, Camera-LiDAR frameworks enhance cooperative perception with camera support~\cite{xiang2023hm}.
While these methods demonstrate excellence in cooperative perception, they often overlook the models' processing capabilities for individual tasks. 
In contrast, our proposed approach not only excels in cooperative perception but also ensures outstanding performance in individual perception.

\section{Methodology}
\label{sec:method}
This section outlines the major components of the SiCP (Simultaneous Individual and Cooperative Perception) framework, illustrated in Figure~\ref{fig:SiCP}.
The framework comprises three key elements: a \textit{feature extractor}, a \textit{Dual-Perception Network}, and an \textit{unified detection head}.
We adopt the backbone of PointPillars~\cite{lang2019pointpillars} as our \textit{feature extractor}, aiming for an optimal balance between effectiveness and efficiency in 3D object detection.
We propose the Dual-Perception Network (DP-Net) that  seamlessly integrates with the existing backbone (Section~\ref{sec:DP-Net}), adeptly processing local features for individual perception and fusing features from multiple vehicles for cooperative perception.  
We employ a unified detection head (Section~\ref{sec:head}) to process features for both individual and cooperative perception tasks.
The proposed solution operates on the premise of vehicular trustworthiness, with all vehicles employing the same machine-learning model for executing their object detection tasks.

\subsection{Dual-Perception Network (DP-Net)}
\label{sec:DP-Net}
The proposed DP-Net module ensures that the individual perception process and cooperative perception process run in parallel.
Moreover, the resulting feature maps, whether for individual or cooperative perception, should be compatible with each other and capable of being processed by a unified detection head.
Specifically, DP-Net executes operations based on whether the ego vehicle has received features from neighboring vehicles.
In the absence of received features, DP-Net continues utilizing current ego vehicle's feature maps, thereby guaranteeing its individual perception performance.
Upon receiving features from neighboring vehicles, DP-Net performs a perspective transformation on these features before forwarding them to the fusion module for integration.
We will delve into the critical endeavor of effectively fusing feature maps in Sections~\ref{sec:feature-sharing} and~\ref{sec:feature-fusion}.
\subsubsection{Receiver-Agnostic Feature Sharing}
\label{sec:feature-sharing}
To enable cooperative perception through deep fusion, vehicles must share their locally generated features with nearby vehicles. 
Achieving efficient vehicular communications requires implementing a receiver-agnostic feature-sharing approach, wherein the sender does not need to know the location and pose of the potential receivers. 
However, implementation of existing methods often require transforming LiDAR data or feature maps into the perspective of the receiving vehicle before transmission~\cite{xu2022opv2v, xu2022v2x, yang2023spatio, wang2023core, qiao2023adaptive}.
In scenarios involving multiple receiving vehicles, this results in creating and transmitting multiple versions of the same feature maps, leading to significant network traffic and computational demands on the sending vehicle.
Alternatively, the sending vehicle can optimize its communication by broadcasting its local feature map to all nearby vehicles.
\begin{figure*}[htbp]
    \centerline{
    \includegraphics[width=6.5in, height=1.6in]{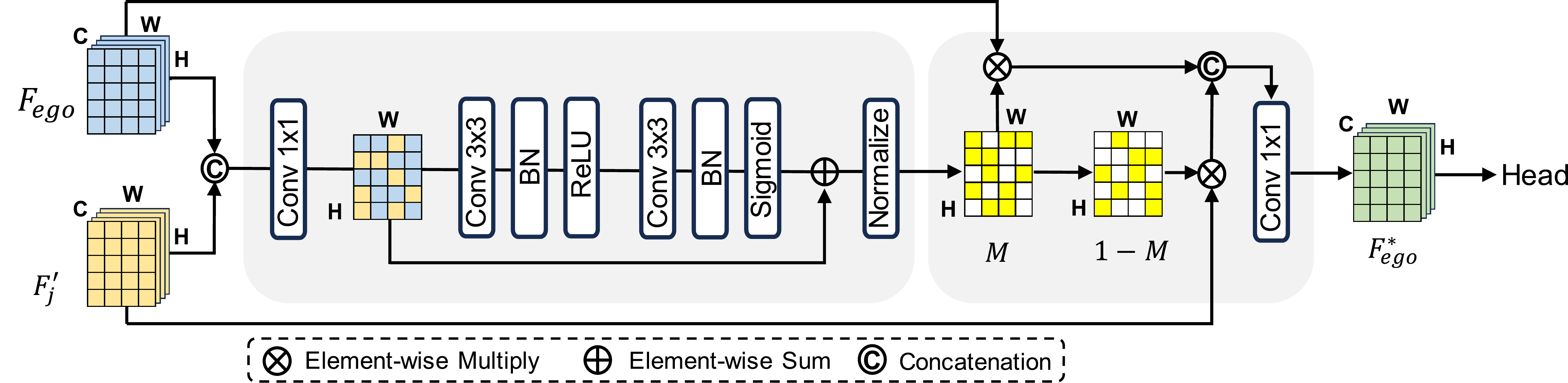}\\
    }
    \caption{Complementary Feature Fusion efficiently merges two BEV (Bird's Eye View) feature maps by learning a weighted map. Initially, it concatenates the two feature maps and condenses them into a one-channel feature map, using a 1x1 convolutional operation. This resultant feature map undergoes processing through two convolutional layers, generating the weighted map $M$. $M$ adjusts the ego vehicle’s local feature map, whereas the complementary weighted map $(1-M)$ modifies the received feature map. Finally, the two feature maps are concatenated and reshaped to the size of $H \times W \times C$.}
    \label{fig:cf-net}
    \vspace{-2mm}
\end{figure*}
In this approach, the sending vehicle shares not just its feature map but also its current location and pose information. 
Upon receiving a shared feature map, the recipient vehicle performs feature transformation using the Affine Transformation technique~\cite{lu2023robust}.

\subsubsection{Complementary Feature Fusion}
\label{sec:feature-fusion}
An essential element of the DP-Net is a fusion module that efficiently merges the received feature map with the locally generated one on the ego vehicle. 
This fusion mechanism depends on preserving gradients within the feature maps earmarked for integration. 

\begin{figure}[!htp]
  \begin{center}
  \includegraphics[width=2.5in, height=1.5in]{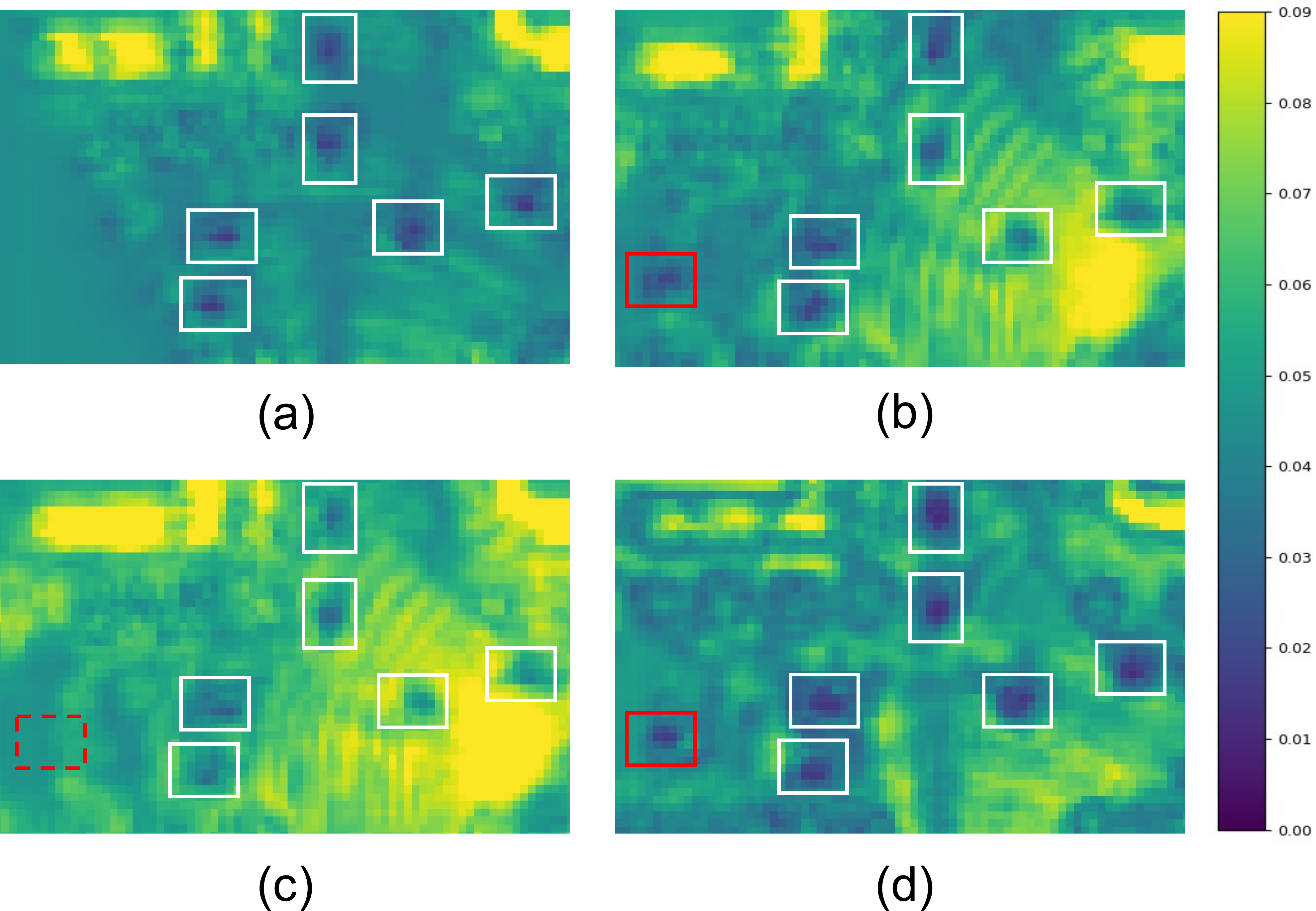}\\
   \caption{Gradient information is lost during the fusion of feature maps. In (a), a receiver's feature map clearly indicates six objects but misinterprets one object (red rectangle, as shown in (b)) in the sender's feature map. Upon fusion of these maps using the \textit{maxout} function, the gradients of this particular object (red dotted rectangle) vanish, as shown in (c). Our method can effectively preserve the gradient during the feature fusion process, as shown in (d). }
   \vspace{-6mm}
    \label{fig:gradients}
    \vspace{2mm}
  \end{center}
\end{figure}

\textbf{Gradients Matter in Fusion.}
Our study has uncovered a useful insight into the features associated with vehicle objects.
As depicted in Figure~\ref{fig:gradients}, features located near the edges of a vehicle exhibit distinct characteristics, however, the central regions of the object lack distinctive features. 
This occurrence can be attributed to the scarcity or absence of LiDAR-generated points within the internal empty spaces of a vehicle. 
In contrast, the vehicle's body effectively reflects LiDAR signals, generating strong features.

Differences in the LiDAR point cloud data acquired by different vehicles can result in varying features for the same object/vehicle.
Due to occlusions, the receiver has difficulty in capturing meaningful features for one object, indicated by the red box in Figure~\ref{fig:gradients} (b).
Consequently, the resulting feature maps might misinterpret these regions as background, resulting in relatively larger numerical representations. 
When the feature maps from the sender and receiver are fused using F-Cooper~\cite{chen2019f} method (\textit{maxout} function), the gradients in the sender's features vanish due to the influence of the larger numbers in the receiver's feature map.

\textbf{Gradients Perseverance in Fusion.}
Expanding on the insights derived from the above analysis, we present a novel fusion module.  
This fusion process is named as \textit{complementary feature fusion}, as illustrated in Figure~\ref{fig:cf-net}. 
The pipeline is designed not only to prioritize the retention of gradients during the fusion process but also to harness the complementary data contributed by other vehicles. 

Let's assume the ego vehicle receives a feature map $\mathcal{F}_{j}$ from another automated vehicle $j$, along with its pose $\mathcal{P}_{j}$ and location $\mathcal{L}_{j}$ information. 
Based on the Affine Transformation $\Psi{(\cdot)}$, we warp the sender's feature map to the ego-vehicle's perspective to get the transformed feature map $\mathcal{F}_{j}^{'} = \Psi{(\mathcal{F}_{j})}$.
We concatenate the ego vehicle's feature map $\mathcal{F}_{ego}$ and the transformed $\mathcal{F}_{j}^{'}$.
The concatenated feature map is then processed by a $1 \times 1$ convolutional layer.
This process can be summarized as 
\begin{equation*}
    \Phi =   Conv (\mathcal{F}_{ego}  \parallel \mathcal{F}_{j}^{'}) \in \mathbb{R}^{H \times W} \tag{1}
\end{equation*}
where $\Phi$ is the resulting feature map, $\parallel$ denotes the concatenation operation.
Here, we utilize a 1x1 convolutional operation to handle stacked feature maps and produce a unified single-channel feature map. 
This step can be substituted with similar operations such as maxout or averaging operations.

To make use of the aggregated feature map, we follow it with a second operation to fully capture spatial dependencies. 
Specifically, we employ a the following operation to get a weight map $\tilde{M}\in \mathbb{R}^{H \times W}$:
\begin{equation*}
    \tilde{M} =\Phi \oplus \left( \sigma \left( BN \left( Conv\left( \delta \left( BN \left(  Conv \left( \Phi \right) \right) \right)\right)\right)\right)\right) \tag{2}
\end{equation*}
where \textit{BN} denotes the Batch Normalization~\cite{ioffe2015batch}, $\delta$ and $\sigma$ refer to the ReLU and Sigmoid functions~\cite{nwankpa2018activation}, and $\oplus$ denotes element-wise summation.
Here, we apply two layers of $3 \times 3$ convolutional operations. 

Next, we normalize $\tilde{M}$ to get a normalized weighted map $M$, in which all the numbers range within $[0, 1]$.
As $M$ is used to weigh and fuse features from two feature maps, we adjust $M$ as follows. 
For any element $m_{ij}$ in $M$, we have
\begin{equation*}
     m_{ij} =
     \begin{cases}
     m_{ij}, & m_{ij} \in \mathcal{F}_{ego} \cap \mathcal{F}_{j}^{'}\\
     0, & otherwise
     \end{cases} \tag{3}
\end{equation*}
As such, the weight map $M$ only provides clues to fuse features within the overlapping area between $\mathcal{F}_{ego}$ and $\mathcal{F}_{j}^{'}$.
While for the non-overlapping area, the weight map is always $0$.
This implies the ego vehicle does not consider other vehicles' data but only relies on its own to detect objects.

After fusing with the received feature map, the ego vehicle's feature map will be updated to $\mathcal{F}_{ego}^{*}\in \mathbb{R}^{C \times H \times W}$ by the following operations:
\begin{equation*}
 \mathcal{F}_{ego}^{*} = Conv \left(\left({M} \otimes \mathcal{F}_{ego} \right) \parallel \left( \left(1 - {M} \right) \otimes\ \mathcal{F}_{j}^{'} \right)\right)   \tag{4}
\end{equation*}
where $\otimes$ denotes the element-wise multiplication. 
Note that the weight map $M$ is responsible for adjusting the ego vehicle's location feature map, while the complementary weight map $(1-M)$ modifies the received feature map. 
This means, in the fused feature map, each point is strongly influenced either by the ego vehicle's feature or by the other vehicle's feature, but not both simultaneously.

\subsection{Detection Head}
\label{sec:head}
The DP-Net we propose generates two potential outcomes: the feature map of the ego vehicle and the fused feature map. 
To maintain compatibility with existing detection models, it is crucial to establish a unified detection head capable of efficiently handling both types of feature maps. 
In this context, we employ a single detection head to manage both individual and cooperative perception, ensuring that the loss functions for each scenario share a consistent format. 
The loss function utilized in our model aligns with the one employed in the PointPillars model~\cite{lang2019pointpillars}.

To realized effective training, we need to ensure an equitable distribution of training data for individual and cooperative perception tasks. 
%
To conserve computational resources, conducting additional training rounds is unfeasible.
Both individual and cooperative perception components in the proposed model must be trained using a single input data and labels.
Therefore, the total number of training rounds remains constant. 
Specifically, our methodology involves inputting a single data instance, encompassing the raw local LiDAR data and a feature map shared by another vehicle, into the network. 
Then, backpropagations are applied, allowing the individual perception and cooperative perception pipelines to be trained jointly.
%

\section{Experiments}
\label{sec:experiments}

\textbf{Datasets.}
We conduct our evaluations on two extensively used datasets: V2V4Real~\cite{xu2023v2v4real} and OPV2V~\cite{xu2022opv2v}. Both supporting Vehicle-to-Vehicle (V2V) cooperative perception research by providing numerous annotated scenes to facilitate algorithm development and evaluation.
\textit{V2V4Real} is a large-scale real-world dataset, collected by two vehicles simultaneously in the same location, providing multi-view sensor datastream.
\textit{OPV2V} is a simulation dataset featuring a variety of virtual cities and environments, co-simulated through OpenCDA~\cite{xu2021opencda} and CARLA~\cite{carla}. 

In accordance with the particular training and testing specifications of our model, we implement a First-Come-First-Serve policy for each frame in both datasets. This approach facilitates the utilization of data from two vehicles: the ego vehicle and the sender vehicle.

\noindent\textbf{Training.}
Our SiCP model enhances learning efficiency through joint training of individual and cooperative perception. 
The gradients obtained from these two tasks are backpropagated sequentially to update the parameters of the backbone, DP-Net, and the detection head. 
Due to the equal amount of training data and identical backbone parameters, the entire network receives an equivalent amount of training, regarding individual and cooperative perceptions. 
This balanced training approach equips our model to effectively handle SiCP tasks, even when dealing with limited training data. 
It's worth noting that our end-to-end training method eliminates the need for pre-training any parameters.
We train the model with one Nvidia RTX 3090 GPU and employ the Adam optimizer~\cite{kingma2014adam} with a learning rate of 0.001 and a batch size of 1 in our model training process.

\noindent\textbf{Inference.}
At the inference stage, following~\cite{xu2023v2v4real,xu2022opv2v}, we use the Average Precision (AP) metric to evaluate the performance of all models on V2V4Real testset and OPV2V testsets (Default and Culver). 
The evaluation used Intersection over Union (IoU) thresholds of 0.5 and 0.7, respectively.
%

\noindent\textbf{Baselines.}
%
We construct baseline models specifically for individual and cooperative perception for comparison with the proposed SiCP method.
In the individual perception task, the baseline model is no fusion, employing the advanced standalone 3D detection network PointPillars~\cite{lang2019pointpillars}, alongside several SOTA deep fusion models: F-Cooper~\cite{chen2019f}, AttFuse~\cite{xu2022opv2v}, V2X-ViT~\cite{xu2022v2x} and CoBEVT~\cite{xu2022cobevt}.
In the cooperative perception task, the baseline model is late fusion,  combining final prediction outputs from multiple vehicles at a later stage, and the same deep fusion models as in the individual perception task.
Notably, SiCP differs from established deep fusion baseline models, where vehicles must first project their respective point clouds onto the ego vehicle's viewpoint; instead, our implementation requires each vehicle to process the data exclusively from its own perspective, reflecting real-world data sharing scenarios between connected vehicles.

\subsection{Quantitative Evaluations}
In the comparative analysis, we evaluate our model against existing benchmarks on two datasets from two different perspectives: \textit{individual perception} and \textit{cooperative perception}.
Our findings reveal that SiCP outperforms other cooperative solutions, meanwhile, SiCP demonstrates satisfactory performance for individual perception task.

\textbf{Evaluation on V2V4Real dataset.}
In cooperative perception scenarios, our SiCP method demonstrates exceptional performance on the V2V4Real Dataset, as depicted in Table~\ref{tab:v2v4real}. 
In a synchronization setting, where vehicles share perception data instantaneously, SiCP dominates with an impressive AP at an IoU=0.7. 
Notably, it surpasses CoBEVT~\cite{xu2022cobevt} by a significant margin of 9.5\% in the critical 0-30m range at this IoU, showcasing its strong, accurate, and reliable near-range detection capabilities crucial for ensuring safety in dense traffic environments.
This improvement is attributed to the incorporation of the complementary feature fusion,
\begin{table}[h]
\centering
\caption{Individual and Cooperative Perception Evaluation on V2V4Real Dataset.}
\label{tab:v2v4real}
\begin{tabularx}{\columnwidth}{@{}l|>{\centering\arraybackslash}X>{\centering\arraybackslash}X>{\centering\arraybackslash}X>{\centering\arraybackslash}X@{}}
\toprule
\multicolumn{1}{l|}{\multirow{2}{*}{\textbf{Method}}} & \multicolumn{4}{c}{{\cellcolor{lightblue}}\textbf{Individual Perception} (AP@IoU=0.5/0.7)} \\ 
 & overall & 0-30m & 30-50m & 50-100m \\ 
\midrule
F-Cooper$\,_{\text{No Fusion}}$~\cite{chen2019f} & 35.4/17.6 & 58.0/29.1 & 23.1/12.3 & 5.6/2.8 \\
AttFuse$\,_{\text{No Fusion}}$~\cite{xu2022opv2v} & 30.2/15.1 & 51.2/25.9 & 19.8/10.7 & 4.4/2.2\\
V2X-ViT$\,_{\text{No Fusion}}$~\cite{xu2022v2x} & 35.3/16.5 & 57.0/26.3 & 22.2/11.5 & 5.1/2.7\\
CoBEVT$\,_{\text{No Fusion}}$~\cite{xu2022cobevt} & 37.7/20.5 & 58.1/34.2 & 22.4/10.8 & 4.9/2.7\\
PointPillars~\cite{lang2019pointpillars} &\textbf{38.6/23.3}  &\textbf{62.5/38.9}  &\textbf{25.3/14.9}  &\textbf{5.7/3.2}  \\
SiCP$\,_{\text{No Fusion}}$(\textbf{Ours}) & 38.0/22.3 & 60.4/37.9 & 24.2/12.4 & 4.1/2.0\\  
\cmidrule(){1-5} 
\multicolumn{1}{l|}{\multirow{2}{*}{\textbf{Method}}} & \multicolumn{4}{c}{{\cellcolor{lightpink}}\textbf{Cooperative Perception} (AP@IoU=0.5/0.7)} \\ 
 & overall & 0-30m & 30-50m & 50-100m \\ 
\midrule
Late Fusion & 42.2/24.1 & 67.4/32.6  & 25.3/10.9 & 11.7/6.7 \\
F-Cooper~\cite{chen2019f} & 47.7/20.2 & 67.9/31.7 & 31.0/13.6 & 23.5/7.2 \\
AttFuse~\cite{xu2022opv2v} & 43.0/18.9 & 61.5/30.3 & 29.1/12.5 & 16.9/5.3 \\
V2X-ViT~\cite{xu2022v2x} & 49.2/20.2 & 70.7/30.3 & 31.0/14.2 & \textbf{26.9}/8.5 \\
CoBEVT~\cite{xu2022cobevt} & \textbf{51.0}/23.0 & 68.4/34.2 & 34.9/15.1 & 26.7/8.8 \\
SiCP (\textbf{Ours}) & 50.0/\textbf{25.6} & \textbf{71.6/43.7} & \textbf{35.8/16.0} & 26.5/\textbf{8.9} \\
\bottomrule
\end{tabularx}
\end{table}
which carefully merges two feature maps by learning optimized weights for every position within the overlapping feature map.
The weights assigned to the ego vehicle and sender vehicle are mutually complementary, aligning seamlessly with the inherent logic of fusing feature maps.

In individual perception scenarios, where vehicles operate without shared data from other vehicles, SiCP still maintains a strong presence, particularly in near and mid-range detection.
Our SiCP solution showcases a significant improvement, with its performance closely approaching that of the advanced standalone model PointPillars~\cite{lang2019pointpillars}.

\textbf{Evaluation on OPV2V dataset.}
As shown in Table~\ref{tab:opv2v}, our SiCP surpasses competing methods with a standout 71.89\% AP at IoU=0.7, reflecting a 1.75\% improvement over the second-ranked V2X-ViT~\cite{xu2022v2x} in cooperative perception scenarios. Particularly in the Culver test set, SiCP secures a 63.02\% AP at IoU=0.7, outpacing V2X-ViT by a substantial 3.26\% thanks to its innovative approach in fusing and interpreting shared data among vehicles.

In individual perception scenarios, SiCP also exhibits strong performance, closely matching the top standalone model, PointPillars~\cite{lang2019pointpillars}. 
However, in the Default test set, CoBEVT~\cite{xu2022cobevt} and AttFuse~\cite{xu2022opv2v} drops by 4.97\% and 7\%, Moreover, in the Culver test set, CoBEVT~\cite{xu2022cobevt} sees a further drop of 8.1\% at IoU=0.7.
This underscores the significance of considering individual perception in the design of cooperative perception models. 
The strength of our model stems from its specialized dedicated pipeline for individual perception, utilizing features derived solely from the ego vehicle.

\begin{table}[h]
\centering
\caption{Individual and Cooperative Perception Evaluation on OPV2V Dataset.}
\label{tab:opv2v}
\begin{tabularx}{\columnwidth}{@{}l|>{\centering\arraybackslash}X>{\centering\arraybackslash}X>{\centering\arraybackslash}X>{\centering\arraybackslash}X@{}}
\toprule
\multicolumn{1}{l|}{\multirow{2}{*}{\textbf{Method}}} & \multicolumn{4}{c}{{\cellcolor{lightblue}}\textbf{Individual Perception} (AP@IoU=0.5/0.7)} \\ 
& \multicolumn{2}{c} {Default test set}   & \multicolumn{2}{c}{Culver test set}\\ 
\midrule
F-Cooper$\,_{\text{No Fusion}}$~\cite{chen2019f} & 68.79 & 56.76 & 75.31 & 59.82 \\
AttFuse$\,_{\text{No Fusion}}$~\cite{xu2022opv2v} & 67.94 & 53.89 & 73.61 & 57.64\\
V2X-ViT$\,_{\text{No Fusion}}$~\cite{xu2022v2x} 
& 72.26 & 57.09 & 75.19 & 56.52\\
CoBEVT$\,_{\text{No Fusion}}$~\cite{xu2022cobevt} & 70.36 & 55.92 & 70.99 & 54.11\\
PointPillars~\cite{lang2019pointpillars} &\textbf{73.84} & \textbf{60.89} & \textbf{77.71} & \textbf{62.21}\\
SiCP$\,_{\text{No Fusion}}$(\textbf{Ours}) & 73.72 & \textbf{60.89} & 76.36 & 61.28\\  
\cmidrule(){1-5} 
\multicolumn{1}{l|}{\multirow{2}{*}{\textbf{Method}}} & \multicolumn{4}{c}{{\cellcolor{lightpink}}\textbf{Cooperative Perception} (AP@IoU=0.5/0.7)} \\ 
& \multicolumn{2}{c} {Default test set}   & \multicolumn{2}{c}{Culver test set}\\ 
\midrule
Late Fusion & 77.03 & 62.95 & 75.59 & 58.45 \\
F-Cooper~\cite{chen2019f} & 79.71 & 63.95 & 68.26 & 48.86 \\
AttFuse~\cite{xu2022opv2v} & 81.91 & 67.00 & 78.00 & 58.11 \\
V2X-ViT~\cite{xu2022v2x} & 85.62 & 70.14 & 79.09 & 59.76 \\
CoBEVT~\cite{xu2022cobevt} & 84.77 & 68.11 & 75.22 & 55.42 \\
SiCP (\textbf{Ours}) &\textbf{85.64} & \textbf{71.89} & \textbf{79.10} & \textbf{63.02} \\
\bottomrule
\end{tabularx}
\end{table}
\subsection{DP-Net is a Robust Module on Alignment Error}
\begin{figure}[!htp]
  \centering
  \begin{subfigure}{.5\columnwidth} 
    \centering
    \includegraphics[width=\linewidth]{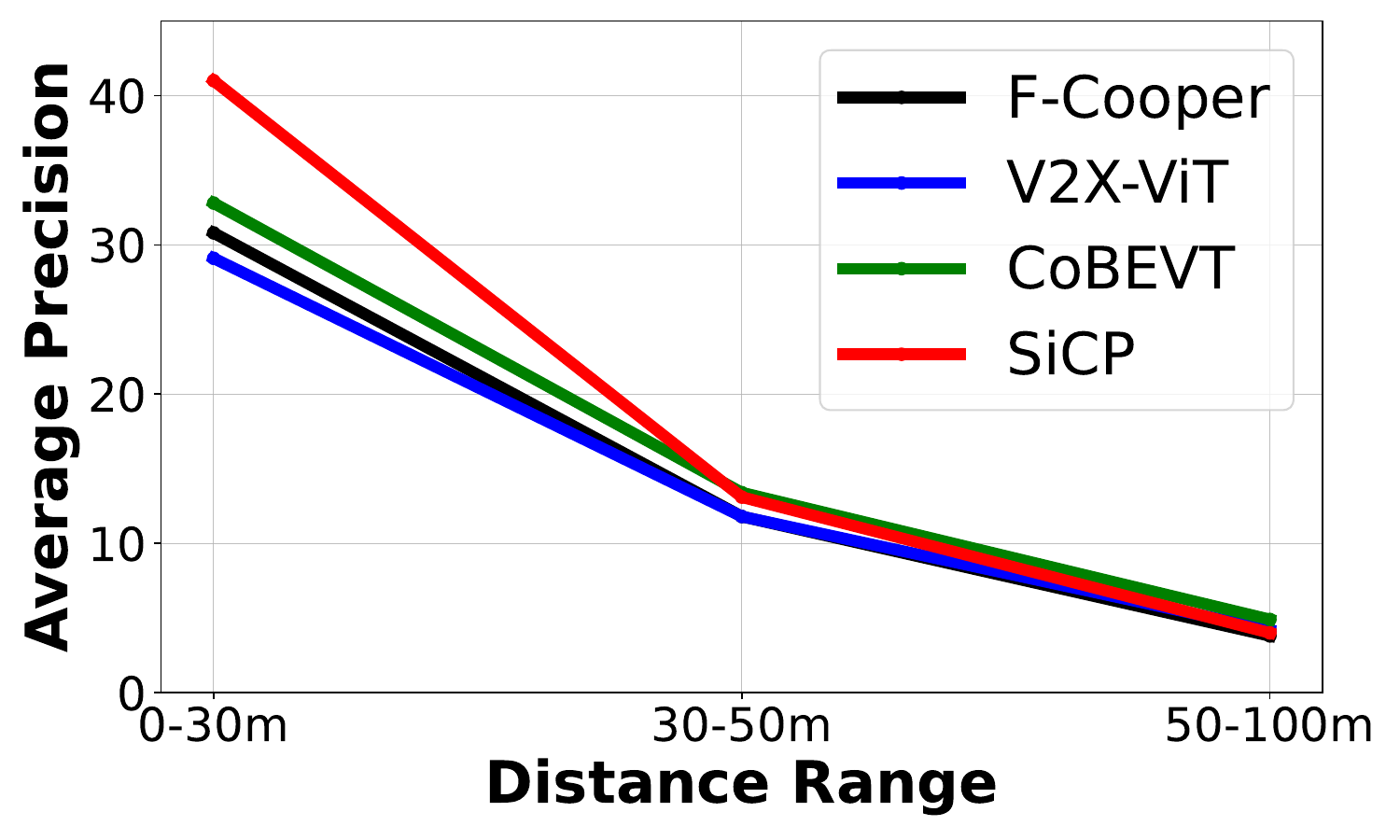} 
    \caption{Asynchronous on V2V4Real}
    \label{fig:sub1}
  \end{subfigure}%
  \begin{subfigure}{.5\columnwidth}
    \centering
    \includegraphics[width=\linewidth]{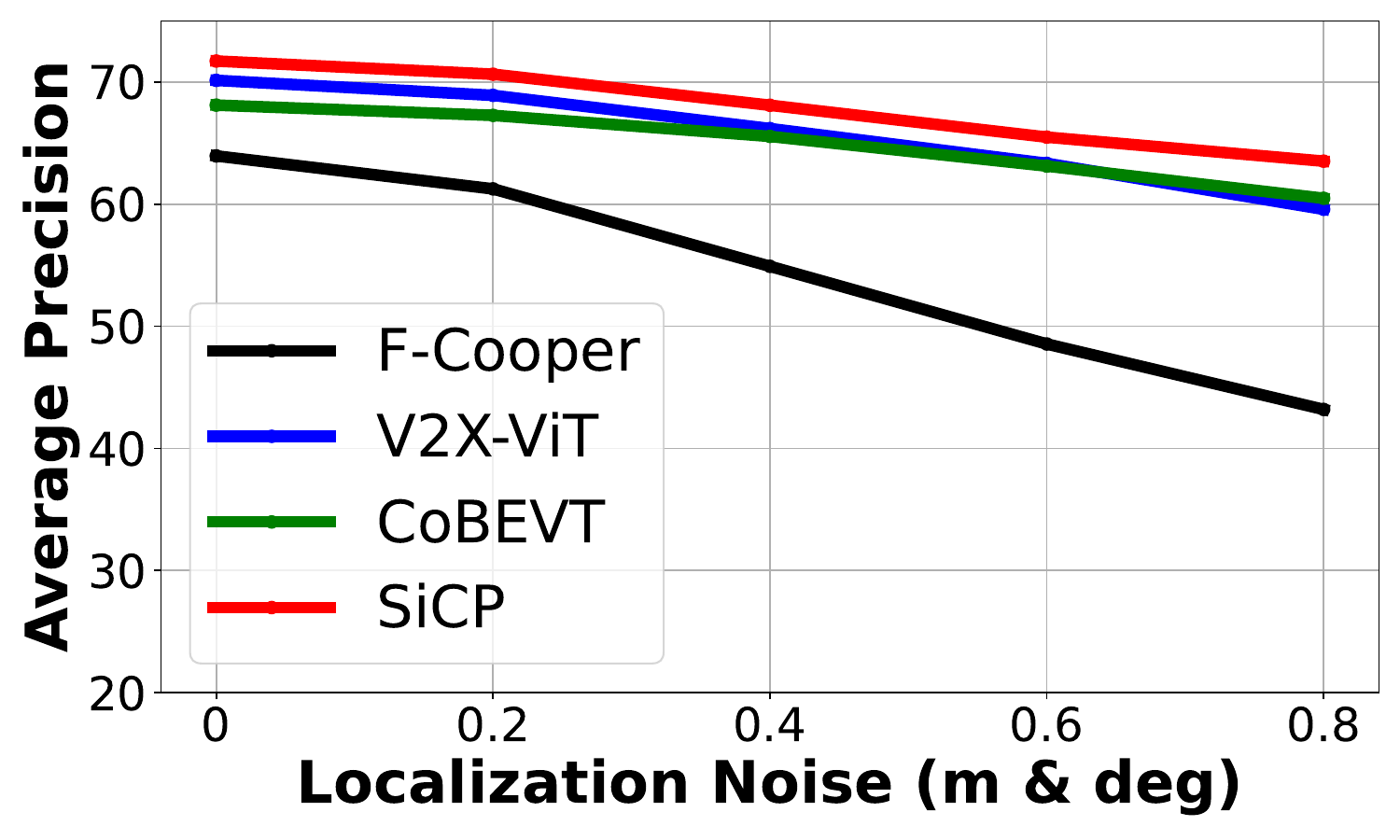} 
    \caption{Loc. error on OPV2V}
    \label{fig:sub2}
  \end{subfigure}
  \caption{Robust response to asynchronous mode and localization error. \textit{SiCP} outperforms other SOTAs across both datasets with IoU=0.7.}
\vspace{-6mm}
  \label{fig:loc_err}
  \vspace{2mm}
\end{figure}

The DP-Net effectively handles alignment errors arising from \textbf{asynchronous communication} and \textbf{inaccurate localization} between vehicles.
As shown in Figure~\ref{fig:loc_err} (a), adapting to an asynchronous environment with a 100 ms communication delay, the resilience of SiCP becomes evident. 
In the 0-30m range, it maintains robust performance at IoU=0.7 with an AP of 41\%, solidifying its ability to process delayed data efficiently. 
SiCP's AP exceeds that of its closest competitor by 8.2\%, underscoring its exceptional skill in near-range detection despite the challenges posed by asynchronous communication.
Furthermore, SiCP demonstrates exceptional robustness to real-world challenges, specifically localization errors between vehicles. 
Through experiments that introduce Gaussian noise to simulate the errors (the noise is set to x, y location and yaw angle), as illustrated in Figure~\ref{fig:loc_err} (b), SiCP consistently maintains its superior performance. 
This showcases its ability to effectively manage inaccuracies in vehicle positioning, underscoring its practical utility and robustness in cooperative perception scenarios.

\subsection{DP-Net is a Lightweight Plug-and-Play Module}
We conduct a comparative analysis involving two pairs of comparisons, each comprising an original backbone model and the one extended by integrating DP-Net. 
The selected standalone 3D detection backbones are two representative network architectures in the field of 3D object detection: PointPillars~\cite{lang2019pointpillars} and VoxelNet~\cite{zhou2018voxelnet}.
As demonstrated in Table~\ref{tab:generic}, DP-Net consistently demonstrates improvement across all baseline 3D detection backbones. 
These findings underscore the generic applicability of DP-Net, suggesting its potential to be integrated into other 3D object detection frameworks.
Moreover, the proposed DP-Net is a lightweight solution as it does not significantly increase the number of parameters requiring training, i.e., DP-Net consists of only $0.13M$ parameters.
If the underlying backbone is PointPillars, with a set of $7.27M$ parameters to be trained, 
\begin{table}[htbp]
\centering
\caption{Existing 3D detection backbones can be extended with DP-Net to address cooperative perception.}
\label{tab:generic}
\begin{tabular}{@{}c|c|cccc@{}}
\toprule
\multicolumn{1}{l|}{\multirow{2}{*}{\textbf{Backbone}}} & \multicolumn{1}{l|}{\multirow{2}{*}{\textbf{Method}}} & \multicolumn{2}{c}{Default AP@IoU} & \multicolumn{2}{c}{Culver AP@IoU} \\ 
\multicolumn{1}{c|}{} & \multicolumn{1}{c|}{} & 0.5 & 0.7 & 0.5 & 0.7 \\ 
\cmidrule(){1-6} 
\multirow{2}{*}{PointPillars~\cite{lang2019pointpillars}} & - & 73.84 & 60.89 & 77.71 & 62.21 \\
 & \textbf{+DP-Net} & \textbf{85.64} & \textbf{71.89} & \textbf{79.10} & \textbf{63.02} \\
\cmidrule(){1-6} 
\multirow{2}{*}{VoxelNet~\cite{zhou2018voxelnet}} & - & 72.99 & 62.13 & 71.86 & 59.95 \\
 & \textbf{+DP-Net} & \textbf{83.23} & \textbf{69.36} & \textbf{77.38} & \textbf{62.73} \\
\bottomrule
\end{tabular}
\end{table}
the introduction of DP-Net increases the overall parameters by a mere $1.7\%$.
With its lightweight design, SiCP achieves a latency of just $37.84 ms$.

\subsection{Qualitative Evaluations}
\begin{figure*}[!htp]
  \begin{center}
  \includegraphics[width=5.9in, height=1.9in]{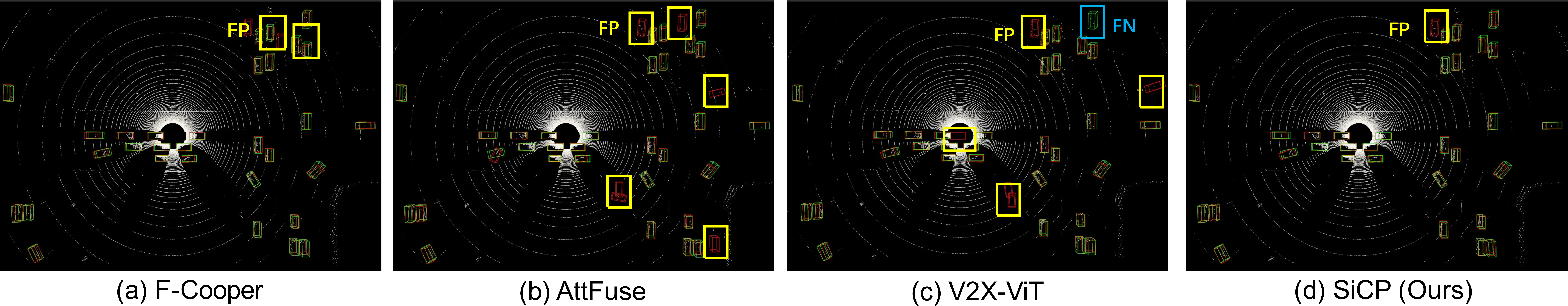}\\
   \caption{Illustrations of \textcolor{orange}{false positive} and \textcolor{cyan}{false negative} in individual and cooperative perception on the OPV2V dataset.}
    \label{fig:qal-opv2v}
  \end{center}
  \vspace{-2mm}
\end{figure*}

\begin{figure*}[!htp]
  \begin{center}
  \includegraphics[width=6.9in, height=1.9in]{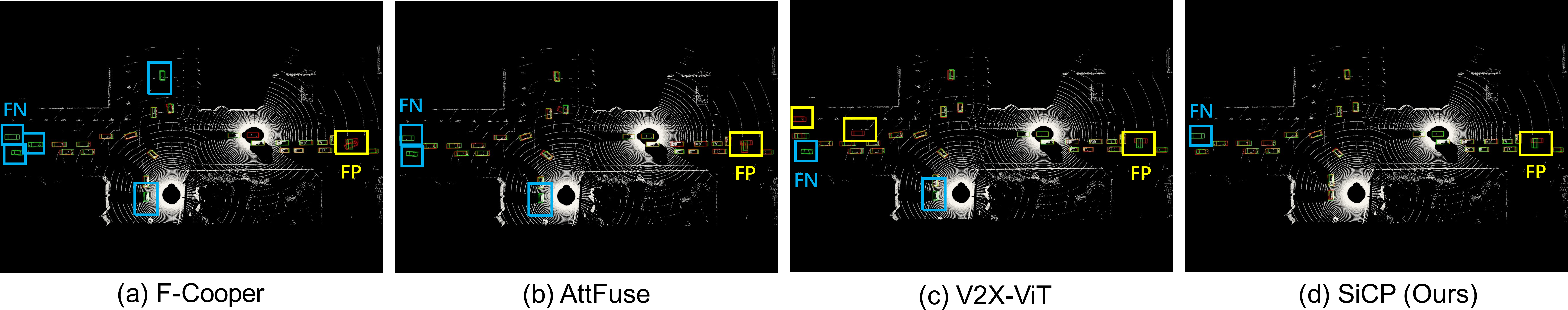}\\
   \caption{Illustrations of \textcolor{orange}{false positive} and \textcolor{cyan}{false negative} in individual and cooperative perception on the V2V4Real dataset.}
    \label{fig:qal-v2v4real}
  \end{center}
  \vspace{-7mm}
\end{figure*}

More qualitative results for individual perception and cooperative perception are presented in both datasets, shown in Figure~\ref{fig:qal-opv2v} and Figure~\ref{fig:qal-v2v4real}. 
Specifically, (a) showcases detection errors for F-Cooper, (b) illustrates errors for V2X-ViT, (c) displays errors for CoBEVT, and (d) demonstrates errors of SiCP. 
Regarding individual perception, both figures highlights instances of false positive and false negative detection results, attributed to incorrect feature extraction in existing cooperative solutions methods while attempting to address the individual perception task. 
In contrast, the SiCP method incorporates both individual and cooperative perception during feature extraction, leading to more precise object detection results.

Regarding cooperative perception, our SiCP model also exhibits significantly fewer false negative and false positive detection results in comparison to (a) F-Cooper, (b) V2X-ViT, and (c) CoBEVT. 
Notably, we observe a higher incidence of false negative detections in cooperative perception, compared to the individual perception. 
This is because any erroneous fusion can generate features that fail to indicate objects accurately, thereby causing false negatives.
The superior performance of SiCP is attributed to DP-Net, which effectively fuses features and generates a more precise representation of objects within the feature maps. 
%

\subsection{Ablation Study}
To better understand how SiCP can simultaneously handle individual and cooperative perception tasks, we conducted ablation studies on the individual perception pipeline and the complementary  weight map. 
As evident from the results in Table~\ref{tab:ablation}, the individual perception pipeline is crucial for ensuring SiCP's capability in individual perception tasks. 
Meanwhile, the complementary weight map highlights the importance of weighting feature maps in a complementary manner, crucial for effective feature fusion. 
Together, these components contribute to SiCP's robustness in addressing various perception challenges.

\begin{table}[htbp]
\centering
\caption{Ablation Study on the Individual Perception Pipeline and Complementary Weight Map ($1-M$) within the Cooperative Perception Pipeline. Results are reported in AP@IoU=0.7 on OPV2V dataset.}
\label{tab:ablation}
\begin{tabular}{@{}cc|c|c@{}}
\toprule
\textbf{Individual} & \textbf{Complementary} & \multicolumn{1}{>{\columncolor{lightblue}}c|}{\textbf{Individual}} & \multicolumn{1}{>{\columncolor{lightpink}}c}{\textbf{Cooperative}} \\ 
\textbf{Perception Pipeline} & \textbf{Weight Map} & Default & Default \\ 
\midrule
 & \checkmark & 31.14 & \textbf{72.00}\\
\midrule
 \checkmark &  & 61.01 & 69.41\\
\midrule
\checkmark & \checkmark & \textbf{63.02} & 71.89\\
\bottomrule
\end{tabular}
  \vspace{-4mm}
\end{table}

\section{Conclusions}
\label{sec:conclusion}
In conclusion, this study shows how to realize simultaneous individual and cooperative perception for connected and automated vehicles. 
Our research demonstrates the feasibility of handling both tasks within a single model, leading to reduced memory usage in automated vehicles and minimized overall interference time. 
This solution is highly practical, given car manufacturers' reluctance to implement separate models for individual and cooperative perception due to the associated high costs. 
The proposed DP-Net is versatile and can be seamlessly incorporated into other standalone 3D object detection models, empowering them with the capability of cooperative perception.

\section{Acknowledgments}
The work is supported by the National Science Foundation grants CNS-2231519, OAC-2017564, and ECCS-2010332.
\bibliographystyle{IEEEtran}
\bibliography{references}

\begin{thebibliography}{10}
\providecommand{\url}[1]{#1}
\csname url@rmstyle\endcsname
\providecommand{\newblock}{\relax}
\providecommand{\bibinfo}[2]{#2}
\providecommand\BIBentrySTDinterwordspacing{\spaceskip=0pt\relax}
\providecommand\BIBentryALTinterwordstretchfactor{4}
\providecommand\BIBentryALTinterwordspacing{\spaceskip=\fontdimen2\font plus
\BIBentryALTinterwordstretchfactor\fontdimen3\font minus \fontdimen4\font\relax}
\providecommand\BIBforeignlanguage[2]{{%
\expandafter\ifx\csname l@#1\endcsname\relax
\typeout{** WARNING: IEEEtran.bst: No hyphenation pattern has been}%
\typeout{** loaded for the language `#1'. Using the pattern for}%
\typeout{** the default language instead.}%
\else
\language=\csname l@#1\endcsname
\fi
#2}}

\bibitem{lang2019pointpillars}
A.~H. Lang, S.~Vora, H.~Caesar, L.~Zhou, J.~Yang, and O.~Beijbom, ``Pointpillars: Fast encoders for object detection from point clouds,'' in \emph{Proceedings of the IEEE/CVF conference on computer vision and pattern recognition}, 2019, pp. 12\,697--12\,705.

\bibitem{yan2018second}
Y.~Yan, Y.~Mao, and B.~Li, ``Second: Sparsely embedded convolutional detection,'' \emph{Sensors}, vol.~18, no.~10, p. 3337, 2018.

\bibitem{zhou2018voxelnet}
Y.~Zhou and O.~Tuzel, ``Voxelnet: End-to-end learning for point cloud based 3d object detection,'' in \emph{Proceedings of the IEEE conference on computer vision and pattern recognition}, 2018, pp. 4490--4499.

\bibitem{chen2019f}
Q.~Chen, X.~Ma, S.~Tang, J.~Guo, Q.~Yang, and S.~Fu, ``F-cooper: Feature based cooperative perception for autonomous vehicle edge computing system using 3d point clouds,'' in \emph{Proceedings of the 4th ACM/IEEE Symposium on Edge Computing}, 2019, pp. 88--100.

\bibitem{xu2022opv2v}
R.~Xu, H.~Xiang, X.~Xia, X.~Han, J.~Li, and J.~Ma, ``Opv2v: An open benchmark dataset and fusion pipeline for perception with vehicle-to-vehicle communication,'' in \emph{2022 International Conference on Robotics and Automation (ICRA)}.\hskip 1em plus 0.5em minus 0.4em\relax IEEE, 2022, pp. 2583--2589.

\bibitem{xu2022v2x}
R.~Xu, H.~Xiang, Z.~Tu, X.~Xia, M.-H. Yang, and J.~Ma, ``V2x-vit: Vehicle-to-everything cooperative perception with vision transformer,'' in \emph{European conference on computer vision}.\hskip 1em plus 0.5em minus 0.4em\relax Springer, 2022, pp. 107--124.

\bibitem{xu2022cobevt}
R.~Xu, Z.~Tu, H.~Xiang, W.~Shao, B.~Zhou, and J.~Ma, ``Cobevt: Cooperative bird's eye view semantic segmentation with sparse transformers,'' \emph{arXiv preprint arXiv:2207.02202}, 2022.

\bibitem{qi2017pointnet}
C.~R. Qi, H.~Su, K.~Mo, and L.~J. Guibas, ``Pointnet: Deep learning on point sets for 3d classification and segmentation,'' in \emph{Proceedings of the IEEE conference on computer vision and pattern recognition}, 2017, pp. 652--660.

\bibitem{shi2019pointrcnn}
S.~Shi, X.~Wang, and H.~Li, ``Pointrcnn: 3d object proposal generation and detection from point cloud,'' in \emph{Proceedings of the IEEE/CVF conference on computer vision and pattern recognition}, 2019, pp. 770--779.

\bibitem{qi2017pointnet++}
C.~R. Qi, L.~Yi, H.~Su, and L.~J. Guibas, ``Pointnet++: Deep hierarchical feature learning on point sets in a metric space,'' \emph{Advances in neural information processing systems}, vol.~30, 2017.

\bibitem{chen2019cooper}
Q.~Chen, S.~Tang, Q.~Yang, and S.~Fu, ``Cooper: Cooperative perception for connected autonomous vehicles based on 3d point clouds,'' in \emph{2019 IEEE 39th International Conference on Distributed Computing Systems (ICDCS)}.\hskip 1em plus 0.5em minus 0.4em\relax IEEE, 2019, pp. 514--524.

\bibitem{arnold2020cooperative}
E.~Arnold, M.~Dianati, R.~de~Temple, and S.~Fallah, ``Cooperative perception for 3d object detection in driving scenarios using infrastructure sensors,'' \emph{IEEE Transactions on Intelligent Transportation Systems}, vol.~23, no.~3, pp. 1852--1864, 2020.

\bibitem{hu2022where2comm}
Y.~Hu, S.~Fang, Z.~Lei, Y.~Zhong, and S.~Chen, ``Where2comm: Communication-efficient collaborative perception via spatial confidence maps,'' \emph{Advances in neural information processing systems}, vol.~35, pp. 4874--4886, 2022.

\bibitem{lu2023robust}
Y.~Lu, Q.~Li, B.~Liu, M.~Dianati, C.~Feng, S.~Chen, and Y.~Wang, ``Robust collaborative 3d object detection in presence of pose errors,'' in \emph{2023 IEEE International Conference on Robotics and Automation (ICRA)}.\hskip 1em plus 0.5em minus 0.4em\relax IEEE, 2023, pp. 4812--4818.

\bibitem{guo2021coff}
J.~Guo, D.~Carrillo, S.~Tang, Q.~Chen, Q.~Yang, S.~Fu, X.~Wang, N.~Wang, and P.~Palacharla, ``Coff: Cooperative spatial feature fusion for 3-d object detection on autonomous vehicles,'' \emph{IEEE Internet of Things Journal}, vol.~8, no.~14, pp. 11\,078--11\,087, 2021.

\bibitem{vnet}
T.-H. Wang, S.~Manivasagam, M.~Liang, B.~Yang, W.~Zeng, and R.~Urtasun, ``V2vnet: Vehicle-to-vehicle communication for joint perception and prediction,'' in \emph{European Conference on Computer Vision}.\hskip 1em plus 0.5em minus 0.4em\relax Springer, 2020, pp. 605--621.

\bibitem{li2021learning}
Y.~Li, S.~Ren, P.~Wu, S.~Chen, C.~Feng, and W.~Zhang, ``Learning distilled collaboration graph for multi-agent perception,'' \emph{Advances in Neural Information Processing Systems}, vol.~34, pp. 29\,541--29\,552, 2021.

\bibitem{xiang2023hm}
H.~Xiang, R.~Xu, and J.~Ma, ``Hm-vit: Hetero-modal vehicle-to-vehicle cooperative perception with vision transformer,'' \emph{arXiv preprint arXiv:2304.10628}, 2023.

\bibitem{ma2024macp}
Y.~Ma, J.~Lu, C.~Cui, S.~Zhao, X.~Cao, W.~Ye, and Z.~Wang, ``Macp: Efficient model adaptation for cooperative perception,'' in \emph{Proceedings of the IEEE/CVF Winter Conference on Applications of Computer Vision}, 2024, pp. 3373--3382.

\bibitem{yu2022dair}
H.~Yu, Y.~Luo, M.~Shu, Y.~Huo, Z.~Yang, Y.~Shi, Z.~Guo, H.~Li, X.~Hu, J.~Yuan, \emph{et~al.}, ``Dair-v2x: A large-scale dataset for vehicle-infrastructure cooperative 3d object detection,'' in \emph{Proceedings of the IEEE/CVF Conference on Computer Vision and Pattern Recognition}, 2022, pp. 21\,361--21\,370.

\bibitem{dhakal2023sniffer}
S.~Dhakal, Q.~Chen, D.~Qu, D.~Carillo, Q.~Yang, and S.~Fu, ``Sniffer faster r-cnn: A joint camera-lidar object detection framework with proposal refinement,'' in \emph{2023 IEEE International Conference on Mobility, Operations, Services and Technologies (MOST)}.\hskip 1em plus 0.5em minus 0.4em\relax IEEE, 2023, pp. 1--10.

\bibitem{newdhakal2023sniffer}
S.~Dhakal, D.~Carrillo, D.~Qu, Q.~Yang, and S.~Fu, ``Sniffer faster r-cnn++: An efficient camera-lidar object detector with proposal refinement on fused candidates,'' \emph{Journal on Autonomous Transportation Systems}, 2023.

\bibitem{dhakal2023virtualpainting}
S.~Dhakal, D.~Carrillo, D.~Qu, M.~Nutt, Q.~Yang, and S.~Fu, ``Virtualpainting: Addressing sparsity with virtual points and distance-aware data augmentation for 3d object detection,'' \emph{arXiv preprint arXiv:2312.16141}, 2023.

\bibitem{yang2023spatio}
K.~Yang, D.~Yang, J.~Zhang, M.~Li, Y.~Liu, J.~Liu, H.~Wang, P.~Sun, and L.~Song, ``Spatio-temporal domain awareness for multi-agent collaborative perception,'' in \emph{Proceedings of the IEEE/CVF International Conference on Computer Vision}, 2023, pp. 23\,383--23\,392.

\bibitem{wang2023core}
B.~Wang, L.~Zhang, Z.~Wang, Y.~Zhao, and T.~Zhou, ``Core: Cooperative reconstruction for multi-agent perception,'' in \emph{Proceedings of the IEEE/CVF International Conference on Computer Vision}, 2023, pp. 8710--8720.

\bibitem{qiao2023adaptive}
D.~Qiao and F.~Zulkernine, ``Adaptive feature fusion for cooperative perception using lidar point clouds,'' in \emph{Proceedings of the IEEE/CVF Winter Conference on Applications of Computer Vision}, 2023, pp. 1186--1195.

\bibitem{ioffe2015batch}
S.~Ioffe and C.~Szegedy, ``Batch normalization: Accelerating deep network training by reducing internal covariate shift,'' in \emph{International conference on machine learning}.\hskip 1em plus 0.5em minus 0.4em\relax pmlr, 2015, pp. 448--456.

\bibitem{nwankpa2018activation}
C.~Nwankpa, W.~Ijomah, A.~Gachagan, and S.~Marshall, ``Activation functions: Comparison of trends in practice and research for deep learning,'' \emph{arXiv preprint arXiv:1811.03378}, 2018.

\bibitem{xu2023v2v4real}
R.~Xu, X.~Xia, J.~Li, H.~Li, S.~Zhang, Z.~Tu, Z.~Meng, H.~Xiang, X.~Dong, R.~Song, \emph{et~al.}, ``V2v4real: A real-world large-scale dataset for vehicle-to-vehicle cooperative perception,'' in \emph{Proceedings of the IEEE/CVF Conference on Computer Vision and Pattern Recognition}, 2023, pp. 13\,712--13\,722.

\bibitem{xu2021opencda}
R.~Xu, Y.~Guo, X.~Han, X.~Xia, H.~Xiang, and J.~Ma, ``Opencda: an open cooperative driving automation framework integrated with co-simulation,'' in \emph{2021 IEEE International Intelligent Transportation Systems Conference (ITSC)}.\hskip 1em plus 0.5em minus 0.4em\relax IEEE, 2021, pp. 1155--1162.

\bibitem{carla}
A.~Dosovitskiy, G.~Ros, F.~Codevilla, A.~Lopez, and V.~Koltun, ``Carla: An open urban driving simulator,'' in \emph{Conference on robot learning}.\hskip 1em plus 0.5em minus 0.4em\relax PMLR, 2017, pp. 1--16.

\bibitem{kingma2014adam}
D.~P. Kingma and J.~Ba, ``Adam: A method for stochastic optimization,'' \emph{arXiv preprint arXiv:1412.6980}, 2014.

\end{thebibliography}

\end{document}